\documentclass[letterpaper, 10 pt, conference]{ieeeconf}  

\IEEEoverridecommandlockouts                              

\usepackage{graphicx} 
\usepackage{amsmath} 
\usepackage{amssymb}  
\usepackage{multirow}
\usepackage{booktabs}
\usepackage{hyperref}
\usepackage[table,xcdraw]{xcolor}
\usepackage[normalem]{ulem}
\useunder{\uline}{\ul}{}

\definecolor{first}{HTML}{D3FFDA}
\definecolor{second}{HTML}{FEFFBB}

\title{\LARGE \bf
FlowMoDE: Coarse-to-fine Flow Matching for Detail-aware Sim-to-Real Monocular Depth Estimation
}

\author{Liangjing Shao$^{1,2}$, Wanhao Liu$^{2}$, Jinsong Lin$^1$, Zhiwei Fang$^1$, Hongliang Ren$^{1,2,\#}$
\thanks{$^{1}$ Liangjing Shao, Jinsong Lin, Zhiwei Fang and Hongliang Ren are with Department of Electronic Engineering, The Chinese University of Hong Kong.}%
\thanks{$^{2}$ Liangjing Shao, Wanhao Liu and Hongliang Ren are also with Shenzhen Loop Area Institute.}%
\thanks{$^{\#}$Corresponding Author: {\tt\small hlren@ee.cuhk.edu.hk}}%
}

\begin{document}

\maketitle
\thispagestyle{empty}
\pagestyle{empty}

\begin{abstract}
Monocular depth estimation is a fundamental capability for 3D spatial perception in robotic and intelligent systems. While data‑driven discriminative methods have achieved compelling performance for monocular depth estimation, generative approaches can reconstruct fine structural details with stronger generalization. However, most existing generative diffusion models suffer from low inference efficiency caused by multi‑step sampling, whereas single‑step diffusion models struggle to recover high‑fidelity depth maps. To mitigate these drawbacks, this paper presents FlowMoDE, a coarse‑to‑fine flow‑matching‑based framework for efficient, structure‑aware monocular depth estimation supporting sim‑to‑real and cross‑scene generalization. The core architecture first conducts coarse‑level decoding conditioned on diffused features extracted from a pre‑trained semantic encoder, and subsequently restores fine‑grained structural patterns via expanded decoding layers. Trained exclusively on a single synthetic indoor dataset, our proposed method outperforms prior methods across five real‑world indoor and outdoor datasets, yielding an average 14.1\% reduction in relative error against state‑of‑the‑art works as well as 9.9\% reduction in chamfer distance for point cloud reconstruction. We further validate the generalization capability of FlowMoDE on realistic robotic scenes, with qualitative comparisons against state‑of‑the‑art methods. The project page is \href{https://flowmode-shao.netlify.app/}{https://flowmode-shao.netlify.app/}.

\end{abstract}

\section{INTRODUCTION}
Monocular Depth Estimation (MDE) is a core 3D perception technique that underpins a wide range of robotic downstream tasks, including scene reconstruction, robotic manipulation, and autonomous navigation. \cite{icra1,icra2} Due to its practical significance, numerous MDE methods have been developed, which are generally divided into discriminative and generative paradigms. Despite delivering promising overall depth prediction performance, discriminative methods inherently suffer from inaccurate depth estimation at object boundaries and fine-grained scene details, which severely degrades the geometric consistency of predicted depth maps. In contrast, generative models alleviate such local detail degradation issues yet introduce inevitable limitations: high computational cost and poor global scene contextual information, resulting in constrained overall prediction accuracy and robustness in complex scenarios.

\begin{figure}
    \centering
    \includegraphics[width=\linewidth]{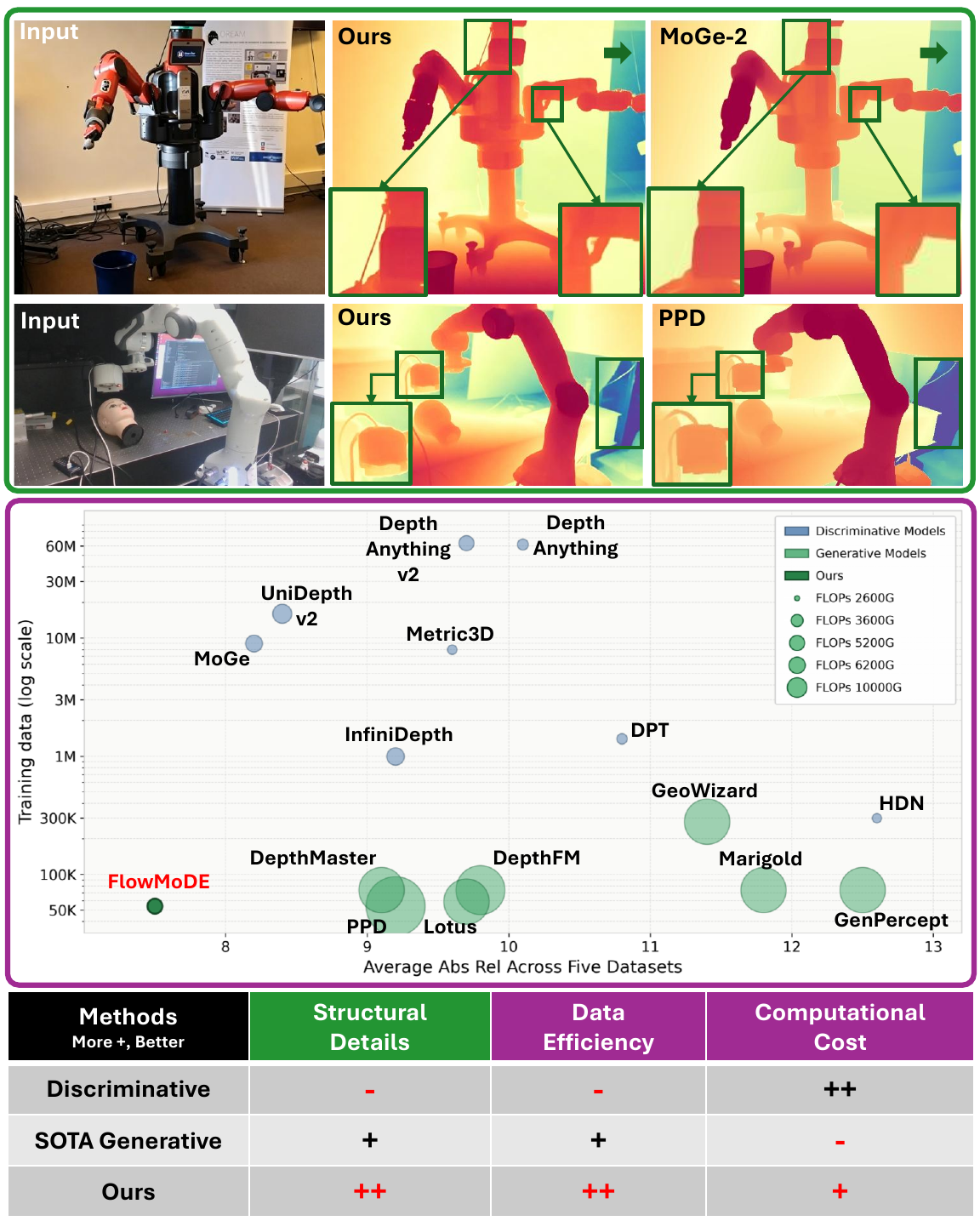}
    \caption{Comparison with existing works. Qualitative comparison in green box demonstrates the \textbf{outstanding structural details} from our method, compared with state-of-the-art discriminative model MoGe-2 and generative model PPD. Bubble graph with purple box denotes comparison of \textbf{performance}, \textbf{data efficiency} and \textbf{computational cost} of the proposed model and existing models.}
    \label{teaser}
\end{figure}

Discriminative architectures \cite{moge,moge2,udepth,udepth2} commonly adopt a large‑patch Vision Transformer encoder, paired with a convolutional up‑sampling decoder. The encoder encodes input images into a low‑resolution token grid and leverages global self‑attention to capture comprehensive semantic information and scene layout. Such tokenization inevitably degrades pixel‑level signals and fine‑grained high‑frequency details in the encoding phase. \cite{ppd} Once these fine details are discarded in early encoding stages, convolutional up‑sampling modules alone cannot completely restore them. As illustrated by the qualitative results within the green box in Fig. \ref{teaser}, this issue yields erroneous depth predictions for fine‑scale structures including wires, holes, and edges of distant objects. Moreover, discriminative depth models typically require large‑scale training datasets covering diverse scenarios to achieve competitive performance. \cite{da,da2} This imposes substantial costs for data collection and hinders their deployment in specialized scenarios where data acquisition is challenging, such as surgical environments. The comparison of data efficiency between generative models and discriminative models is demonstrated in purple box of Fig. \ref{teaser}.

Current generative models \cite{marigold,monodiffusion} commonly fine‑tune Stable Diffusion to perform depth estimation. These approaches rely on a Variational Autoencoder (VAE) to project depth maps into a latent space. Such compression inherently sacrifices edge sharpness and structural fidelity. Therefore, recent studies have attempted to conduct diffusion directly in the pixel space for better recovery of structural details.\cite{ppd} Nevertheless, the multi‑step pixel‑wise diffusion process introduces high computational overhead. While several existing approaches \cite{lotus,depthmaster} adopt single‑step Stable Diffusion for depth estimation, the heavy VAE encoder and decoder still create significant computation bottlenecks. Meanwhile, single-step denoising can limit the performance of depth estimation. 

In this work, we propose FlowMoDE, a coarse‑to‑fine flow matching model for computationally efficient sim‑to‑real pixel‑accurate depth estimation. A lightweight decoder named as Semantic Decoder extracts global semantic features from a pre‑trained discriminative encoder. Guided by global features, our model performs efficient decoding to establish a coarse depth distribution through Coarse Decoder. A subsequent Fine-grained Decoder with few wide Transformers further recovers delicate structural details to produce the final pixel-perfect depth map. Our main contributions can be summarized as the following:
\begin{enumerate}
    \item We propose a novel flow matching framework for sim-to-real monocular depth estimation, which can provide structure-aware depth maps with lower computational cost.
    \item We introduce coarse-to-fine semantic-guided velocity prediction which efficiently extracts coarse features based on semantic prior and further recover delicate details.
    \item The proposed model is only trained on synthetic indoor dataset, but achieve state-of-the-art performance across \textbf{five} diverse datasets including \textbf{both indoor and outdoor scenes}. Further application of the proposed model in robotic scenes demonstrate the generalization and potential ability for spatial perception of robots.
\end{enumerate}

\section{RELATED WORK}

\subsection{Discriminative Monocular Depth Estimation}
Early monocular depth estimation relied on hand‑crafted visual priors and graphical models to infer scene layouts from single images. Deep learning redefined MDE as a supervised dense prediction task, giving rise to various CNN‑based solutions such as \cite{cnn1}. Later relative‑depth studies such as MiDaS \cite{midas} prioritized cross‑dataset generalization, proving that mixing heterogeneous datasets improves zero‑shot transfer. Meanwhile, DPT \cite{dpt} firstly established a high‑performance transformer‑driven dense prediction framework, leading to a series of excellent ViT-based methods. Recent foundation models such as Depth Anything \cite{da,da2} enhance open‑domain robustness via large‑scale training. In parallel, metric‑depth approaches such as Metric3D \cite{metric3d} and UniDepth \cite{udepth,udepth2} aim to reconstruct depth with absolute scale and have achieved steady improvements in accuracy and generalization. Despite these advances, modern discriminative models commonly adopt large‑patch ViT encoders together with convolutional upsampling decoders. Many discriminative methods strengthen detail reconstruction through local refinement, high‑resolution inference or multi‑scale decoding. Depth‑Pro \cite{depthpro} produces sharp metric depth via high‑resolution architectures and boundary‑aware training. MoGe \cite{moge} improves local geometry by predicting affine‑invariant point maps under combined global and local geometric supervision, while MoGe‑2 \cite{moge2} further preserves fine details using refined training data. MDA \cite{mda} models per‑pixel depth ambiguity with mixture‑density representations to generate point maps without flying‑point noise. InfiniDepth \cite{idepth} formulates depth as a neural implicit field and employs local implicit decoders to support arbitrary‑resolution queries and clearer depth details. Nevertheless, early tokenization in such architectures damages pixel‑level cues and high‑frequency components, making fine geometric structures hard to recover in decoding.

\subsection{Generative Monocular Depth Estimation}
Diffusion generative models have been applied in monocular depth estimation models such as Marigold \cite{marigold} and MonoDiffusion \cite{monodiffusion}, benefiting from rapid convergence and strong representation priors. PPD \cite{ppd} introduces the injection of semantic features into late diffusion process to maintain global information and perform pixel-level denoise to provide pixel-perfect depth maps. However, the computational cost acutely increase due to multiple diffusion transformer layers. Recently, DepthMaster \cite{depthmaster} utilize single-step UNet-based diffusion to perform generative depth estimation. However, heavy VAE still limit computational efficiency, while single-step denoise degrades the performance of the proposed model. DepthFM \cite{depthfm} firstly introduces flow matching into monocular depth estimation. However, lack of semantic information limits its performance, while VAE-based architecture still cause high computational cost and loss of structural details. To this end, FlowMoDE proposed in this work injects adjusted semantic features into few-step flow matching to efficiently perform structure-aware monocular depth estimation.

\section{METHOD}

\begin{figure*}[!ht]
    \centering
    \includegraphics[width=\linewidth]{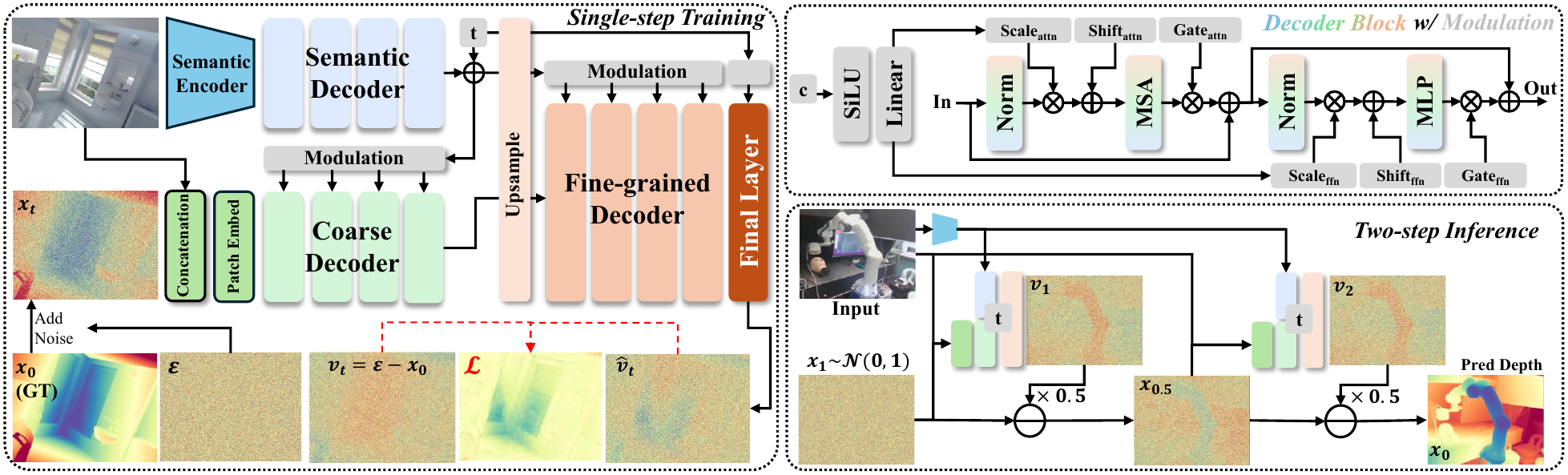}
    \caption{Pipeline of the proposed method.}
    \label{pipe}
\end{figure*}

\subsection{Preliminaries: Flow Matching}
Flow matching learns a time-dependent velocity field $v_\theta(x_t,t)$ to transport samples from a simple source distribution $p_0$ to the data distribution $p_1$ over timestep $t$.
The generative process is governed by the ODE:
\begin{equation}
\frac{dx_t}{dt}=v_\theta(x_t,t),x_0\sim p_0,x_1\sim p_1,t\in[0,1]
\end{equation}
In conditional flow matching, the intermediate point is $x_t=(1-t)x_0+tx_1$.
The training loss minimizes the $\ell_2$ distance between the predicted velocity field and the analytical target $x_1-x_0$:
\begin{equation}
\mathcal{L}_{\text{CFM}}(\theta)=\mathbb{E}\Big[\big\|v_\theta(x_t,t)-(x_1-x_0)\big\|_2^2\Big].
\end{equation}

\subsection{Overview}
The overall pipeline of the proposed method is demonstrated in Fig. \ref{pipe}. Given an input image $I$, the corresponding ground-truth depth map with noise $D_t$ and its semantic feature map $s$, the proposed coarse-to-fine velocity prediction model $\mathcal{F}_{diff}$ predict the velocity map $\hat{v}=\mathcal{F}_{diff}([I,D_t]|s)$ for interpolation matching at the training stage. At the inference stage, given the input image $I$ and random noise $x_1\sim \mathcal{N}(0,1)$, FlowMoDE subsequently utilizes the shared coarse-to-fine velocity prediction model for twice to predict the depth map $\hat{D}$.

\subsection{Coarse-to-Fine Velocity Prediction}
Firstly, shown as Eq.\ref{sem}, semantic feature $s^0$ of the input image $I$ is extracted by a semantic encoder $\mathcal{F}_{sem}$, which can be pretrained models such as DINOv2. 
\begin{equation}
    s^0=Linear_{d_s\rightarrow d_c}(Norm(\mathcal{F}_{sem}(I)))
    \label{sem}
\end{equation}
where $d_s$ is the dimension of the output from semantic encoder, which means $\mathcal{F}_{sem}(I)\in\mathbb{R}^{B\times\frac{H}{P}\times\frac{W}{P}\times d_s}$. Moreover, $d_c$ is the dimension of the condition token, which is set to 384 in this work. The semantic feature $s^0$ is further processed by a \textit{Semantic Decoder} with four Transformer blocks to obtain semantic condition $s$.

The timestep $t$ is sampled from $[0,T=1000]$ based on normal noise $\epsilon_t\sim\mathcal{N}(0,1)$ by equation Eq. \ref{t}. Meanwhile, the ground-truth depth map $D$ is added by noise $\epsilon_d\sim\mathcal{N}(0,1)$ through the process formulated as Eq. \ref{dt} and concatenated with the input image $I$. The concatenated image $I_{cd}=Concat(D_t,I-0.5)$ is then embedded to 16$\times$16 patches as the input of \textit{Coarse Decoder}.
\begin{equation}
    t=round(\frac{T}{1+e^{-\epsilon_t}})
    \label{t}
\end{equation}
\begin{equation}
    D_t=(1-\frac{t}{T})(Norm_d(\log(D+1))-0.5)+\frac{t}{T}\epsilon_d
    \label{dt}
\end{equation}
where $Norm_d$ means linear normalization using 2\%–98\% quantiles.

During training stage, the time condition $t$ is added with the semantic condition $s$, then modulated to factors of scale, shift and gate for multi-head attention $\beta_{m},\gamma_{m},g_{m}$ and MLP in feed forward network $\beta_{f},\gamma_{f},g_{f}$ by SiLU activation and Linear projection layer, formulated as Eq. \ref{mod}.
\begin{equation}
    [\beta_{m},\gamma_{m},g_{m},\beta_{f},\gamma_{f},g_{f}]=Linear_{d_c\rightarrow6d_c}(SiLU(c))
\label{mod}
\end{equation}

The coarse outputs the coarse feature map based on the embedded input through four conditioned Transformer blocks. For each conditioned Transformer block, denoted $x$ as the input and $y$ as the output, the modulated condition is injected into Transformer by Eq. \ref{inj1} and Eq. \ref{inj2} to guide the flow matching process with semantic information.
\begin{equation}
    y_{msa}=x+g_m\cdot Attn((1+\beta_m) \cdot LN(x)+\gamma_m)
\label{inj1}
\end{equation}
\begin{equation}
    y=y_{msa}+g_f\cdot MLP((1+\beta_f)\cdot LN(y_{msa})+\gamma_f)
\label{inj2}
\end{equation}

After coarse decoder, the output feature map is upsampled two times to 8$\times$8 patches. Meanwhile, the condition $c=s+t$ and timestep $t$ are also respectively upsampled to $c_f$ and $t_f$ by Eq.\ref{up}, where $NR$ means repeat the nearest element in $c$ at each position of $c_f$ to upsample $c$.
\begin{equation}
    c_f=Linear_{d_c\rightarrow d_f}(NR(c)), t_f=Linear_{d_c\rightarrow d_f}(t)
    \label{up}
\end{equation}
The fine-grained feature map $f_{fine}$ is obtained from the coarse feature map by four conditioned Transformer block with condition $c_f$. In the final layer, the fine-grained map goes through layer normalization followed by a linear layer with modulation of time token $t_f$, shown as Eq. \ref{final1} and \ref{final2}.
\begin{equation}
    [\beta,\gamma] = Linear_{d_f\rightarrow2d_f}(SiLU(t_f))
    \label{final1}
\end{equation}
\begin{equation}
    f_{final}=Linear_{d_f\rightarrow p^2}((1+\beta)\cdot LN(f_{fine})+\gamma)
    \label{final2}
\end{equation}

At last, we unpatchify the final feature map $f_{final}$ from the final layer to provide the velocity map $\hat{v}$.

\begin{table*}[!ht]
\centering
\caption{Results on Marigold benchmarks. $^*$VAE Encoder \& Decoder are not considered in original reported FLOPs calculation. $^\#$NO reported value, estimated from the description in the original paper. \textbf{Best models} from \textit{ALL models} are in bold, while \underline{Second Best models} are underlined. \colorbox{first}{Best results} and \colorbox{second}{Second Best results} from \textit{Sim-to-Real Generative Models} are marked by corresponding colors.}
\begin{tabular}{lccccccccccccc
>{\columncolor[HTML]{FFFFFF}}c }
\toprule
\multicolumn{1}{c|}{}                                          &                                 &                                                                                    & \multicolumn{1}{c|}{}                                                                               & \multicolumn{2}{c|}{\textbf{NYUv2}}                                                            & \multicolumn{2}{c|}{\textbf{KITTI}}                                                         & \multicolumn{2}{c|}{\textbf{ETH3D}}                                                               & \multicolumn{2}{c|}{\textbf{ScanNet}}                                           & \multicolumn{2}{c|}{\textbf{DIODE}}                                                             & \cellcolor[HTML]{FFFFFF}                                                                              \\ \cline{5-14}
\multicolumn{1}{c|}{\multirow{-2}{*}{\textbf{Method}}}         & \multirow{-2}{*}{\textbf{Year}} & \multirow{-2}{*}{\textbf{\begin{tabular}[c]{@{}c@{}}Training\\ Data\end{tabular}}} & \multicolumn{1}{c|}{\multirow{-2}{*}{\textbf{\begin{tabular}[c]{@{}c@{}}FLOPs\\ (G)\end{tabular}}}} & \textbf{AR$\downarrow$}                         & \multicolumn{1}{c|}{\textbf{$\delta_1$$\uparrow$}}               & \textbf{AR$\downarrow$}                      & \multicolumn{1}{c|}{\textbf{$\delta_1$$\uparrow$}}               & \textbf{AR$\downarrow$}                         & \multicolumn{1}{c|}{\textbf{$\delta_1$$\uparrow$}}                  & \textbf{AR$\downarrow$}                & \multicolumn{1}{c|}{\textbf{$\delta_1$$\uparrow$}}         & \textbf{AR$\downarrow$}                       & \multicolumn{1}{c|}{\textbf{$\delta_1$$\uparrow$}}                  & \multirow{-2}{*}{\cellcolor[HTML]{FFFFFF}\textbf{\begin{tabular}[c]{@{}c@{}}Avg\\ Rank\end{tabular}}} \\ \hline
\multicolumn{15}{c}{\cellcolor[HTML]{EFEFEF}\textit{\textbf{Data-driven Discrimitive Models}}}                                                                                          \\
\multicolumn{1}{l|}{\textbf{MiDaS \cite{midas}}}                            & 2020                            & 2M                                                                                 & \multicolumn{1}{c|}{184.6}                                                                          & 11.1                                 & \multicolumn{1}{c|}{88.5}                               & 23.6                              & \multicolumn{1}{c|}{63.0}                               & 18.4                                 & \multicolumn{1}{c|}{75.2}                                  & 12.1                        & \multicolumn{1}{c|}{84.6}                         & 33.2                               & \multicolumn{1}{c|}{71.5}                                  & 19.2                                                                                                  \\
\multicolumn{1}{l|}{\textbf{DiverseDepth \cite{ddepth}}}                     & 2021                            & 320K                                                                               & \multicolumn{1}{c|}{207.8}                                                                          & 11.7                                 & \multicolumn{1}{c|}{87.5}                               & 19.0                              & \multicolumn{1}{c|}{70.4}                               & 22.8                                 & \multicolumn{1}{c|}{69.4}                                  & 10.9                        & \multicolumn{1}{c|}{88.2}                         & 37.6                               & \multicolumn{1}{c|}{63.1}                                  & 19.6                                                                                                  \\
\multicolumn{1}{l|}{\textbf{LeReS \cite{leres}}}                            & 2021                            & 354K                                                                               & \multicolumn{1}{c|}{190.4}                                                                          & 9.0                                  & \multicolumn{1}{c|}{91.6}                               & 14.9                              & \multicolumn{1}{c|}{78.4}                               & 17.1                                 & \multicolumn{1}{c|}{77.7}                                  & 9.1                         & \multicolumn{1}{c|}{91.7}                         & 27.1                               & \multicolumn{1}{c|}{76.6}                                  & 16.7                                                                                                  \\
\multicolumn{1}{l|}{\textbf{Omnidata \cite{odata}}}                         & 2021                            & 12.2M                                                                              & \multicolumn{1}{c|}{195.4}                                                                          & 7.4                                  & \multicolumn{1}{c|}{94.5}                               & 14.9                              & \multicolumn{1}{c|}{83.5}                               & 16.6                                 & \multicolumn{1}{c|}{77.8}                                  & 7.5                         & \multicolumn{1}{c|}{93.6}                         & 33.9                               & \multicolumn{1}{c|}{74.2}                                  & 16.7                                                                                                  \\
\multicolumn{1}{l|}{\textbf{DPT \cite{dpt}}}                              & 2021                            & 1.4M                                                                               & \multicolumn{1}{c|}{449.2}                                                                          & 9.8                                  & \multicolumn{1}{c|}{90.3}                               & 10.0                              & \multicolumn{1}{c|}{90.1}                               & 7.8                                  & \multicolumn{1}{c|}{94.6}                                  & 8.2                         & \multicolumn{1}{c|}{93.4}                         & \textbf{18.2}                      & \multicolumn{1}{c|}{75.8}                                  & 14.6                                                                                                  \\
\multicolumn{1}{l|}{\textbf{HDN \cite{hdn}}}                              & 2022                            & 300K                                                                               & \multicolumn{1}{c|}{195.4}                                                                          & 6.9                                  & \multicolumn{1}{c|}{94.8}                               & 11.5                              & \multicolumn{1}{c|}{86.7}                               & 12.1                                 & \multicolumn{1}{c|}{83.3}                                  & 8.0                         & \multicolumn{1}{c|}{93.9}                         & 24.6                               & \multicolumn{1}{c|}{78.0}                                  & 14.4                                                                                                  \\
\multicolumn{1}{l|}{\textbf{Metric3D \cite{metric3d}}}                         & 2023                            & 8M                                                                                 & \multicolumn{1}{c|}{225.4}                                                                          & 5.8                                  & \multicolumn{1}{c|}{96.3}                               & \textbf{5.8}                      & \multicolumn{1}{c|}{\textbf{97.0}}                      & 6.6                                  & \multicolumn{1}{c|}{96.0}                                  & 7.4                         & \multicolumn{1}{c|}{94.1}                         & 22.4                               & \multicolumn{1}{c|}{78.5}                                  & 9.6                                                                                                   \\
\multicolumn{1}{l|}{\textbf{DA \cite{da}}}                               & 2024                            & 62M                                                                                & \multicolumn{1}{c|}{586.0}                                                                          & 4.3                                  & \multicolumn{1}{c|}{\textbf{98.1}}                      & 7.6                               & \multicolumn{1}{c|}{94.7}                               & 6.5                                  & \multicolumn{1}{c|}{97.2}                                  & 4.2                         & \multicolumn{1}{c|}{98.0}                         & 27.7                               & \multicolumn{1}{c|}{75.9}                                  & 6.6                                                                                                   \\
\multicolumn{1}{l|}{\textbf{DA V2 \cite{da2}}}                            & 2024                            & 63.5M                                                                              & \multicolumn{1}{c|}{1816.5}                                                                         & 4.4                                  & \multicolumn{1}{c|}{97.9}                               & 7.5                               & \multicolumn{1}{c|}{94.8}                               & 6.2                                  & \multicolumn{1}{c|}{{\ul 98.0}}                            & 4.3                         & \multicolumn{1}{c|}{{\ul 98.1}}                   & 26.0                               & \multicolumn{1}{c|}{75.9}                                  & 6.0                                                                                                   \\
\multicolumn{1}{l|}{\textbf{UniDepth v2 \cite{udepth2}}}                    & 2025                            & 16M                                                                                & \multicolumn{1}{c|}{1925.0}                                                                         & {\ul 4.2}                            & \multicolumn{1}{c|}{{\ul 98.0}}                         & 8.1                               & \multicolumn{1}{c|}{92.6}                               & 4.5                                  & \multicolumn{1}{c|}{97.6}                                  & \textbf{3.7}                & \multicolumn{1}{c|}{\textbf{98.2}}                & 21.6                               & \multicolumn{1}{c|}{78.7}                                  & {\ul 3.7}                                                                                             \\
\multicolumn{1}{l|}{\textbf{MoGe \cite{moge}}}                             & 2025                            & 9M                                                                                 & \multicolumn{1}{c|}{3108.0}                                                                         & 4.4                                  & \multicolumn{1}{c|}{97.9}                               & 7.9                               & \multicolumn{1}{c|}{92.9}                               & 4.8                                  & \multicolumn{1}{c|}{97.3}                                  & {\ul 4.1}                   & \multicolumn{1}{c|}{{\ul 98.1}}                   & 19.7                               & \multicolumn{1}{c|}{{\ul 79.9}}                            & 4.2                                                                                                   \\
\multicolumn{1}{l|}{\textbf{InfiniDepth \cite{idepth}}}                      & 2026                            & $\sim$1M$^\#$                                                                      & \multicolumn{1}{c|}{1844.3}                                                                         & 5.0                                  & \multicolumn{1}{c|}{97.3}                               & 8.7                               & \multicolumn{1}{c|}{91.6}                               & 5.2                                  & \multicolumn{1}{c|}{96.9}                                  & 4.9                         & \multicolumn{1}{c|}{97.3}                         & 22.4                               & \multicolumn{1}{c|}{78.1}                                  & 7.9                                                                                                   \\ \hline
\multicolumn{15}{c}{\cellcolor[HTML]{EFEFEF}\textit{\textbf{Sim-to-Real Generative Models}}}                                                                                                                                                                                                                                                                                                                                                                                                                                                                                                                                                                                                                                                                                                                                                                                               \\
\multicolumn{1}{l|}{\textbf{Marigold \cite{marigold}}}                         & 2024                            & 74K                                                                                & \multicolumn{1}{c|}{5196.5}                                                                         & 5.5                                  & \multicolumn{1}{c|}{96.4}                               & 9.9                               & \multicolumn{1}{c|}{91.6}                               & 6.5                                  & \multicolumn{1}{c|}{96.0}                                  & 6.4                         & \multicolumn{1}{c|}{95.1}                         & 30.8                               & \multicolumn{1}{c|}{77.3}                                  & 12.7                                                                                                  \\
\multicolumn{1}{l|}{\textbf{GeoWizard \cite{geowizard}}}                        & 2025                            & 280K                                                                               & \multicolumn{1}{c|}{5247.3}                                                                         & 5.2                                  & \multicolumn{1}{c|}{96.6}                               & 9.7                               & \multicolumn{1}{c|}{92.1}                               & 6.4                                  & \multicolumn{1}{c|}{96.1}                                  & 6.1                         & \multicolumn{1}{c|}{95.3}                         & 29.7                               & \multicolumn{1}{c|}{\cellcolor[HTML]{FEFFBB}79.2}          & 10.6                                                                                                  \\
\multicolumn{1}{l|}{\textbf{DepthFM \cite{depthfm}}}                          & 2025                            & 74K                                                                                & \multicolumn{1}{c|}{6192.5$^*$}                                                                     & 5.5                                  & \multicolumn{1}{c|}{96.3}                               & 8.9                               & \multicolumn{1}{c|}{91.3}                               & 5.8                                  & \multicolumn{1}{c|}{96.2}                                  & 6.3                         & \multicolumn{1}{c|}{95.4}                         & 22.4                               & \multicolumn{1}{c|}{78.5}                                  & 10.4                                                                                                  \\
\multicolumn{1}{l|}{\textbf{GenPercept \cite{genp}}}                       & 2025                            & 74K                                                                                & \multicolumn{1}{c|}{5237.2$^*$}                                                                     & 5.2                                  & \multicolumn{1}{c|}{96.6}                               & 9.4                               & \multicolumn{1}{c|}{92.3}                               & 6.6                                  & \multicolumn{1}{c|}{95.7}                                  & 5.6                         & \multicolumn{1}{c|}{96.5}                         & 35.7                               & \multicolumn{1}{c|}{75.6}                                  & 12.2                                                                                                  \\
\multicolumn{1}{l|}{\textbf{Lotus \cite{lotus}}}                            & 2025                            & 59K                                                                                & \multicolumn{1}{c|}{5237.2$^*$}                                                                     & 5.4                                  & \multicolumn{1}{c|}{96.8}                               & 8.5                               & \multicolumn{1}{c|}{92.2}                               & 5.9                                  & \multicolumn{1}{c|}{97.0}                                  & 5.9                         & \multicolumn{1}{c|}{95.7}                         & 22.8                               & \multicolumn{1}{c|}{73.8}                                  & 10.2                                                                                                  \\
\multicolumn{1}{l|}{\textbf{DepthMaster \cite{depthmaster}}}                      & 2026                            & 74K                                                                                & \multicolumn{1}{c|}{5241.5$^*$}                                                                     & 5.0                                  & \multicolumn{1}{c|}{97.2}                               & 8.2                               & \multicolumn{1}{c|}{\cellcolor[HTML]{FEFFBB}93.7}       & 5.3                                  & \multicolumn{1}{c|}{97.4}                                  & 5.5                         & \multicolumn{1}{c|}{96.7}                         & \cellcolor[HTML]{FEFFBB}21.5       & \multicolumn{1}{c|}{77.6}                                  & 6.9                                                                                                   \\
\multicolumn{1}{l|}{\textbf{PPD \cite{ppd}}}                              & 2026                            & 54K                                                                                & \multicolumn{1}{c|}{18158.8}                                                                        & \cellcolor[HTML]{FEFFBB}4.3          & \multicolumn{1}{c|}{\cellcolor[HTML]{FEFFBB}97.4}       & \cellcolor[HTML]{FEFFBB}8.0       & \multicolumn{1}{c|}{93.1}                               & \cellcolor[HTML]{FEFFBB}4.5          & \multicolumn{1}{c|}{\cellcolor[HTML]{FEFFBB}97.7}          & \cellcolor[HTML]{FEFFBB}4.5 & \multicolumn{1}{c|}{\cellcolor[HTML]{FEFFBB}97.3} & 24.5                               & \multicolumn{1}{c|}{76.9}                                  & 6.1                                                                                                   \\
\multicolumn{1}{l|}{\cellcolor[HTML]{EFEFEF}\textbf{FlowMoDE}} & \cellcolor[HTML]{EFEFEF}-       & \cellcolor[HTML]{EFEFEF}54K                                                        & \multicolumn{1}{c|}{\cellcolor[HTML]{EFEFEF}3140.1}                                                 & \cellcolor[HTML]{D3FFDA}\textbf{4.0} & \multicolumn{1}{c|}{\cellcolor[HTML]{D3FFDA}\textbf{98.1}} & \cellcolor[HTML]{D3FFDA}{\ul 6.2} & \multicolumn{1}{c|}{\cellcolor[HTML]{D3FFDA}{\ul 95.9}} & \cellcolor[HTML]{D3FFDA}\textbf{3.9} & \multicolumn{1}{c|}{\cellcolor[HTML]{D3FFDA}\textbf{98.8}} & \cellcolor[HTML]{D3FFDA}4.3 & \multicolumn{1}{c|}{\cellcolor[HTML]{D3FFDA}97.6} & \cellcolor[HTML]{D3FFDA}{\ul 18.8} & \multicolumn{1}{c|}{\cellcolor[HTML]{D3FFDA}\textbf{80.4}} & \cellcolor[HTML]{F2F2F2}\textbf{2.2}                                                                  \\ \bottomrule
\end{tabular}
\label{comp}
\end{table*}

\subsection{Training: Single-step Flow Matching}
Given ground-truth (GT) depth map $D$ and noise $\epsilon$, the target velocity map is $v^*=\epsilon-D$. The single coarse-to-fine flow matching model $\mathcal{F}_{diff}$ outputs predicted velocity map $\hat{v}$ from the input image $I$ and noisy GT depth map $D_t$ from Eq. \ref{dt}. We utilize MSE between prediction and target velocity as the loss function for training, formulated as Eq. \ref{loss}.
\begin{equation}
    \mathcal{L}=\frac{\sum_im_i||\hat{v}-v^*||^2_2}{\sum_im_i+\epsilon}
    \label{loss}
\end{equation}
where $m$ denotes the mask of valid depth, where value of invalid area will be set to 0.

\subsection{Two-step Inference}
During training stage, semantic tokens $s$ are extracted once from the input RGB image. Firstly, the shared coarse-to-fine network $\mathcal{F}_{diff}$ predicts velocity map from the concatenation of input RGB image $I$ and random noise $x_1\sim\mathcal{N}(0,1)$. The original noise map is further diffused by the predicted velocity map, formulated as Eq. \ref{s1}.
\begin{equation}
    x_{0.5}=x_1-0.5\times\mathcal{F}_{diff}([x_1, I-0.5]|s)
    \label{s1}
\end{equation}

As Eq. \ref{s2} shows, $x_{0.5}$ concatenated with the input image $I$ is further processed by the shared coarse-to-fine flow matching model to generate the final velocity map and yield the final clean depth map $\hat{D}=x_0+0.5$.

\begin{equation}
    x_0=x_{0.5}-0.5\times\mathcal{F}_{diff}([x_{0.5}, I-0.5]|s)
    \label{s2}
\end{equation}

\begin{figure*}[!ht]
    \centering
    \includegraphics[width=0.9\linewidth]{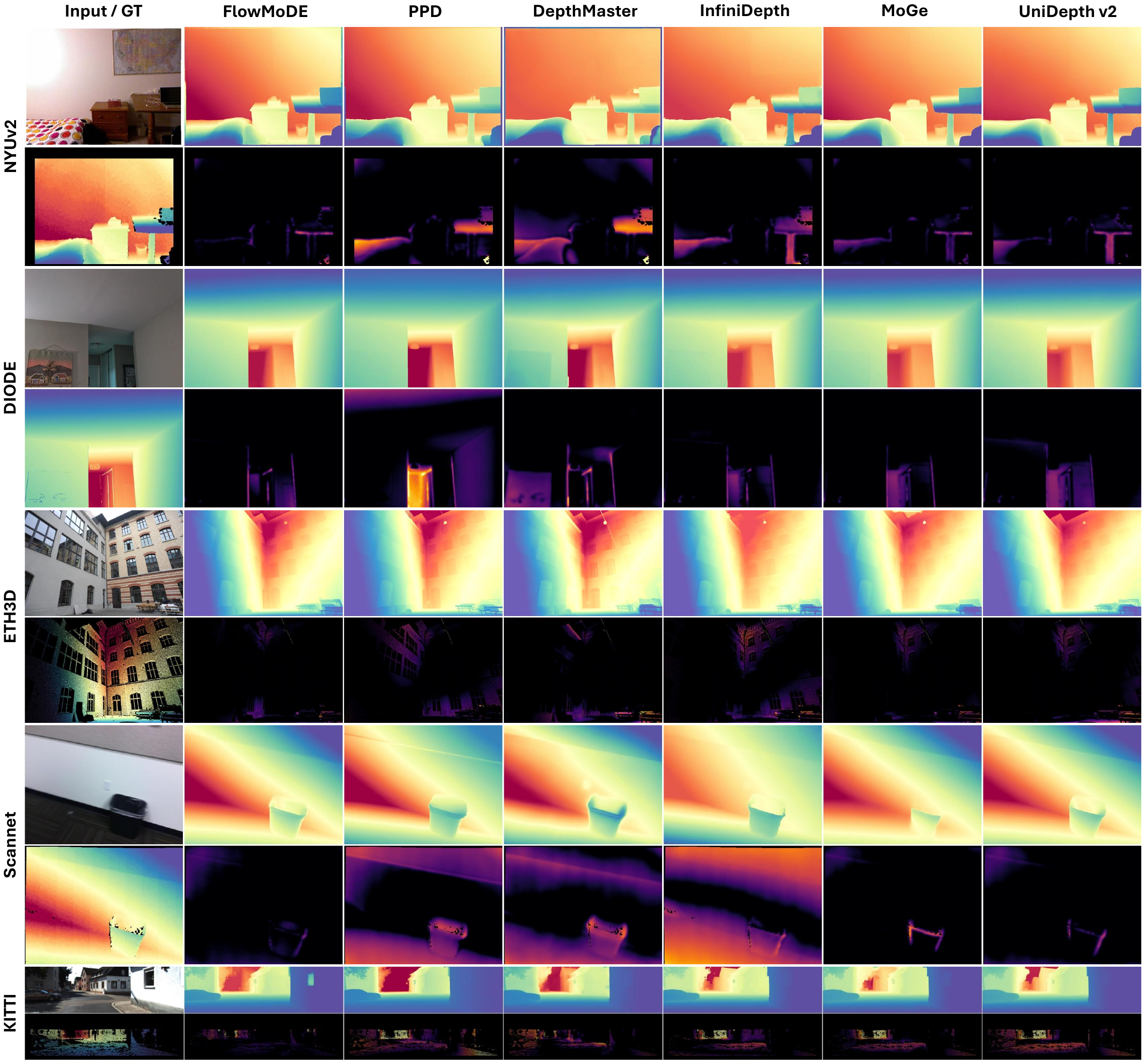}
    \caption{Qualitative comparison on indoor examples (first four rows) and outdoor examples (last four rows). Figures in the second row of each sample are GT and corresponding error maps based on Abs Rel.}
    \label{qual}
\end{figure*}

\section{EXPERIMENTS AND RESULTS}

\subsection{Implementation Details}
We utilize frozen DINOv2-B from MoGe as our semantic encoder. The dimensions of semantic decoder, coarse decoder and fine-grained decoder are respectively set to 384, 384 and 768. We train the proposed model on single dataset Hypersim \cite{hypersim}, which consists of 54K high-quality synthetic indoor frames and corresponding depth maps. The proposed model is trained for 200 epochs (2000 sampling steps for each epoch) with a batch size of 4 and a fixed learning rate of 1e-4, while all training samples are resized to 640$\times$480. All experiments are implemented on a NVIDIA H100 GPU.

\subsection{Comparison Results: Monocular Depth Estimation}
All methods are evaluated on five diverse real-world datasets NYUv2 \cite{nyu}, KITTI \cite{kitti}, ETH3D \cite{eth3d}, ScanNet \cite{scannet} and DIODE \cite{diode} following Marigold \cite{marigold}, resizing all samples to a resolution with a long edge of 768. We compared our model with existing discriminative models and generative models. For DIODE, we directly use masks provided officially, while some works such as PPD \cite{ppd} use Sobel to filter areas with high gradient and RANSAC to align pixels in their papers. We utilize absolute relative error (AB) and accuracy ($\delta$) formulated as Eq.\ref{m1} and Eq. \ref{m2}, where $D$ is the ground truth of the depth map and $\hat{D}$ is the predicted depth map, to evaluate the performance. 
\begin{equation}
    \mathrm{AR}=\frac{1}{|\mathbf{D}|}\sum_{\hat{d}\in\mathbf{D}}|d-\hat{d}|/d
    \label{m1}
\end{equation}
\begin{equation}
    \delta=\frac{1}{|\mathbf{D}|}\left|\left\{\hat{d}\in\mathbf{D}|max(\frac{d}{\hat{d}},\frac{\hat{d}}{d})<1.25\right\}\right|
    \label{m2}
\end{equation}

The quantitative comparison results are shown in Table \ref{comp}. Our model outperforms existing generative models across five datasets, achieving state-of-the-art performance among all compared models with highest average rank on 10 metrics. Displaying results of two different indoor scenes and two different outdoor scenes, Fig. \ref{qual} further demonstrate the outstanding performance of the proposed method.

\begin{table}
\centering
\caption{Quantitative comparison of point cloud reconstruction based on monocular depth estimation. \colorbox{first}{Best results} and \colorbox{second}{Second Best results} are marked by corresponding colors.}
\begin{tabular}{l|cccc|ll}
\hline
\rowcolor[HTML]{C0C0C0} 
\multicolumn{1}{c|}{\cellcolor[HTML]{FFFFFF}}                                  & \multicolumn{2}{c}{\cellcolor[HTML]{C0C0C0}\textbf{Near-field}}       & \multicolumn{2}{c|}{\cellcolor[HTML]{C0C0C0}\textbf{Far-field}} & \multicolumn{2}{c}{\cellcolor[HTML]{EFEFEF}}
                                                                         \\
\rowcolor[HTML]{EFEFEF} 
\multicolumn{1}{c|}{\cellcolor[HTML]{FFFFFF}}                                  & \multicolumn{2}{c}{\cellcolor[HTML]{EFEFEF}\textbf{NYUv2}}            & \multicolumn{2}{c|}{\cellcolor[HTML]{EFEFEF}\textbf{ETH3D}}     & \multicolumn{2}{c}{\multirow{-2}{*}{\cellcolor[HTML]{EFEFEF}\textbf{Mean}}}
                                          \\
\rowcolor[HTML]{FFFFFF} 
\multicolumn{1}{c|}{\multirow{-3}{*}{\cellcolor[HTML]{FFFFFF}\textbf{Method}}} & \textbf{Acc$\downarrow$}                         & \textbf{CD$\downarrow$}                  & \textbf{Acc$\downarrow$}                  & \textbf{CD$\downarrow$}                   & \multicolumn{1}{c}{\cellcolor[HTML]{FFFFFF}\textbf{Acc$\downarrow$}} & \multicolumn{1}{c}{\cellcolor[HTML]{FFFFFF}\textbf{CD$\downarrow$}} \\ \hline
\rowcolor[HTML]{FFFFFF} 
\textbf{UniDepth   v2}                                                         & \cellcolor[HTML]{D3FFDA}86.7          & \cellcolor[HTML]{FEFFBB}117.2 & 332.4                          & 427.1                          & \cellcolor
[HTML]{FEFFBB}209.6                             & 272.2                                                    \\
\rowcolor[HTML]{FFFFFF} 
\textbf{MoGe}
                                                                  & 92.7                                  & 121.4                         & 352.4                          & 448.7                          & 222.6                                                     & 285.1                                                    \\
\rowcolor[HTML]{FFFFFF} 
\textbf{InfiniDepth}                                                           & 100.4                                 & 130.7                         & 352.3                          & 463.3                          & 226.4                                                     & 297.0                                                    \\ \hline
\rowcolor[HTML]{FFFFFF} 
\textbf{DepthMaster}                                                           & 100.0                                 & 120.9                         & \cellcolor[HTML]{FEFFBB}321.8  & \cellcolor[HTML]{FEFFBB}322.3  & 210.9                                                     & \cellcolor
[HTML]{FEFFBB}221.6                            \\
\rowcolor[HTML]{FFFFFF} 
\textbf{PPD}                                                                   & 172.2                                 & 160.3                         & 840.7                          & 596.4                          & 506.5                                                     & 378.4                                                    \\ \hline
\rowcolor[HTML]{D3FFDA} 
\cellcolor[HTML]{F2F2F2}\textbf{FlowMoDE}                                      & \cellcolor[HTML]{FEFFBB}\textbf{89.3} & \textbf{100.2}                & \textbf{269.6}                 & \textbf{299.1}                 & 179.5                                                     & 199.7                                                    \\ \hline
\end{tabular}
\label{rec}
\end{table}

\subsection{Comparison Results: Point Cloud Reconstruction}
To evaluate the downstream application and structural details of monocular depth estimation, we utilize NYUv2 and ETH3D, which provide camera intrinsics for 3D ground-truth, to compare point cloud reconstruction from predicted depth maps. We choose three discriminative models and two generative models with best performance on depth estimation to compare with our model. Accuracy (Acc) and Chamfer Distance (CD) formulated as Eq. \ref{m3} and Eq. \ref{m4} based on point clouds from depth maps (Eq. \ref{pcd}) are utilized as the metrics for point cloud reconstruction.
\begin{equation}
    P(u,v)=K^{-1}[u,v,1]D(u,v)
    \label{pcd}
\end{equation}
\begin{equation}
    Acc=\frac{1}{|\mathcal{P}|}\sum_{p\in \mathcal{P}}\min_{g\in \mathcal{G}}||p-g||_2
    \label{m3}
\end{equation}
\begin{equation}
    CD=\frac{1}{2}(\frac{1}{|\mathcal{P}|}\sum_{p\in \mathcal{P}}\min_{g\in \mathcal{G}}||p-g||_2+ \frac{1}{|\mathcal{G}|}\sum_{g\in \mathcal{G}}\min_{p\in \mathcal{P}}||g-p||_2)
    \label{m4}
\end{equation}
where $\mathcal{P}$ is the point cloud from predicted depth map and $\mathcal{G}$ is the point cloud from ground-truth depth map. For ETH3D dataset, we undistort RGB images and corresponding depth maps following official document \footnote{https://eth3d.ethz.ch/documentation} to match pinhole camera hypothesis before using Eq. \ref{pcd}.

\begin{figure}
    \centering
    \includegraphics[width=\linewidth]{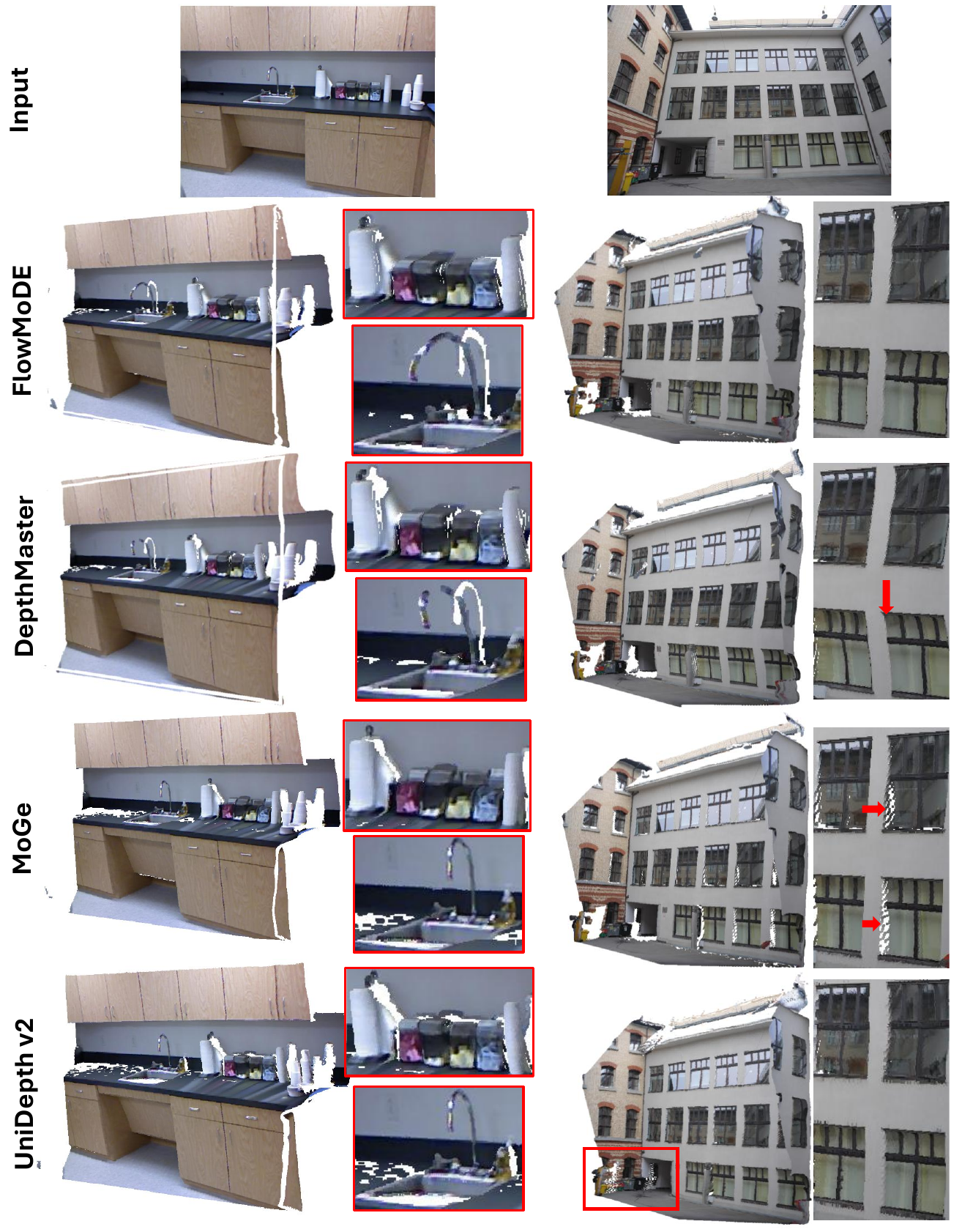}
    \caption{Comparison of point cloud reconstruction based on estimated depth maps.}
    \label{recon}
\end{figure}

Based on results shown in Table. \ref{rec}, our model achieve the outstanding performance with best average metrics across two datasets, while UniDepth v2 can provide results with better accuracy on NYUv2. Furthermore, we perform qualitative comparison in Fig. \ref{recon}. We can find that the proposed model can provide better spatial perception on structural details for delicate objects, such as the faucet in the first column of Fig. \ref{recon} and windows in the second column of Fig. \ref{recon}.

\subsection{Comparison Results: 3D Perception in Robotic Scenes}
\begin{figure}
    \centering
    \includegraphics[width=\linewidth]{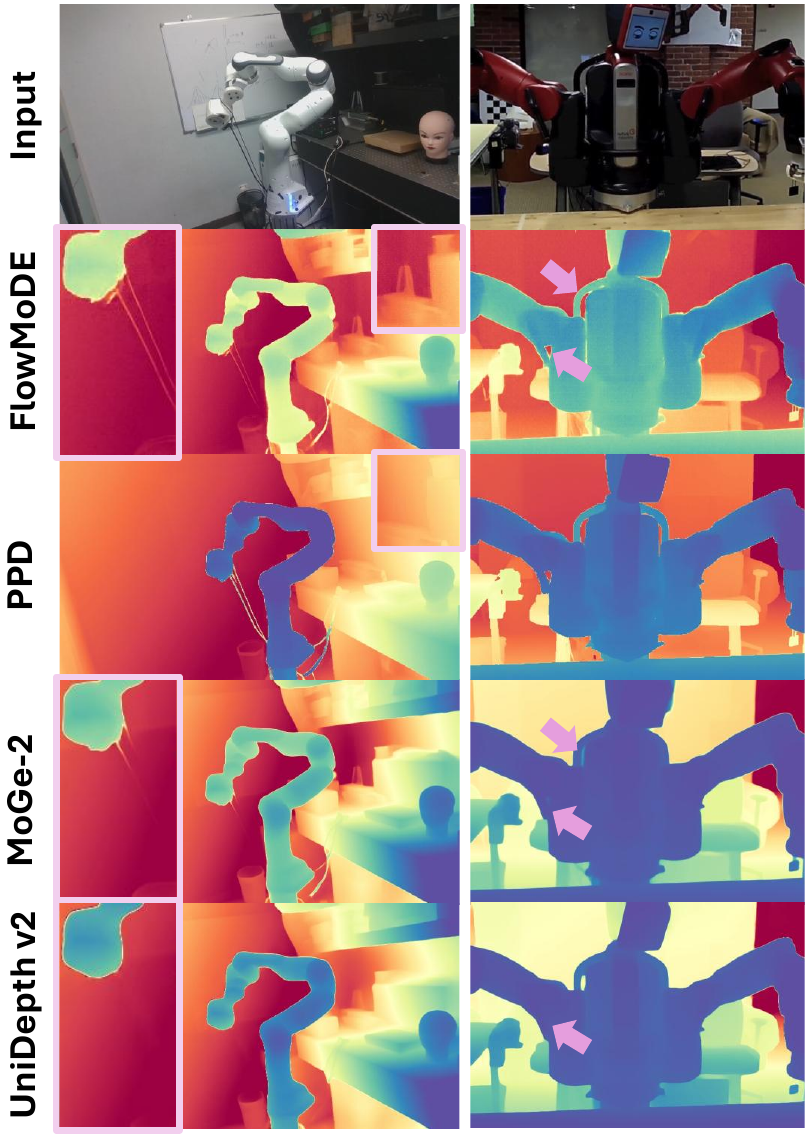}
    \caption{Qualitative comparison of monocular depth estimation in robotic scenarios.}
    \label{rob_dp}
\end{figure}

\begin{figure}
    \centering
    \includegraphics[width=\linewidth]{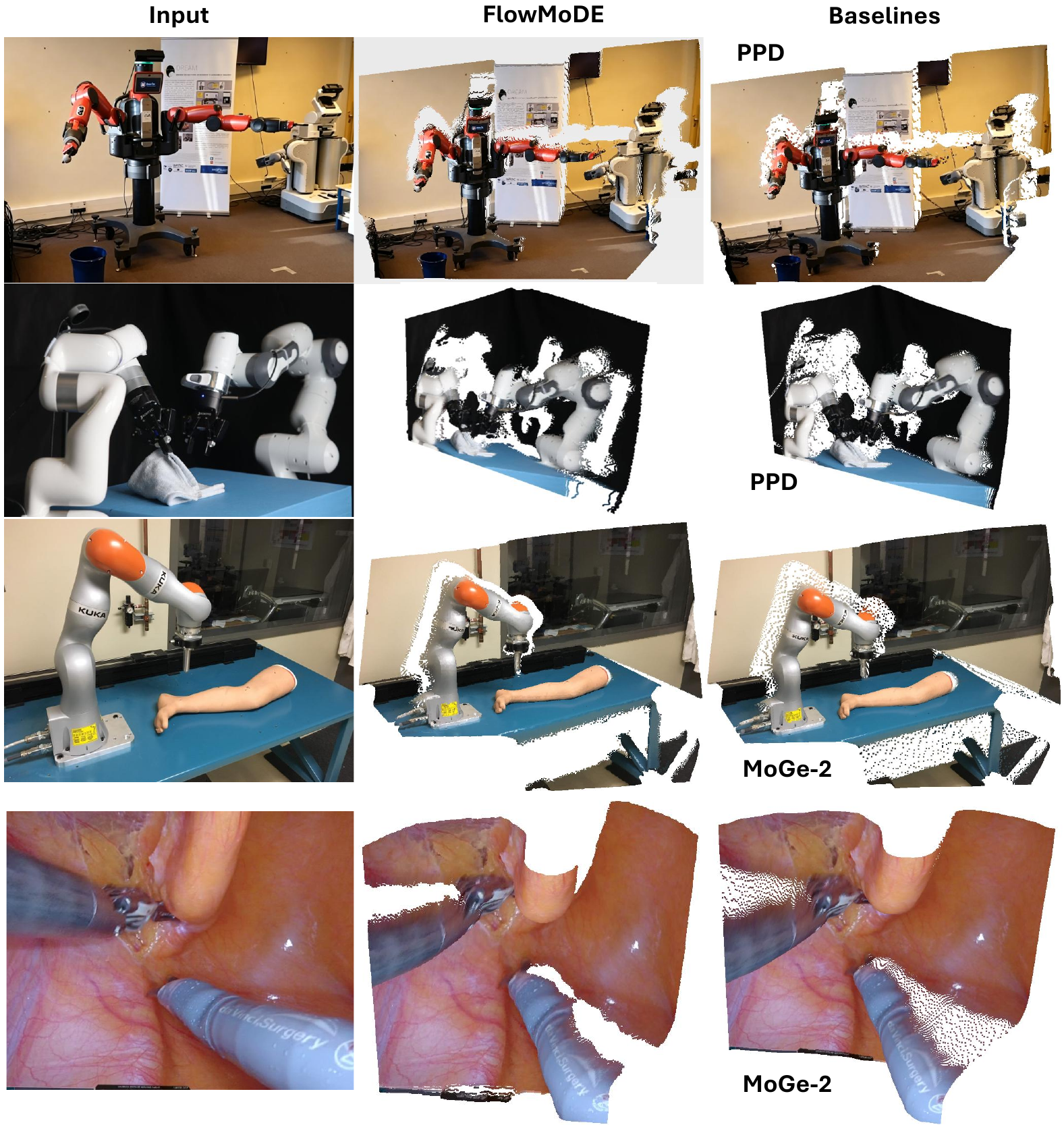}
    \caption{Qualitative comparison of point cloud reconstruction based on estimated depth maps in diverse cross-embodiment robotic scenes.}
    \label{rob}
\end{figure}

In contrast to conventional settings, robotic environments typically contain multiple distinct agents that interact with each other. Therefore, we choose diverse cross-embodiment robotic scenes which respectively consist of Baxter, Franka, Kuka and Da Vinci Robot, to implement the proposed model for zero-shot monocular depth estimation. Compared with the most recent methods in Fig. \ref{rob_dp}, we can find that our proposed method can perceive subtle details such as the delicate structure of robot body or cables. Meanwhile, based on 3D reconstruction results compared with PPD or MoGe-2 in Fig. \ref{rob}, FlowMoDE can better reconstruct the robot body independent of the environment or objects, reducing flying points between them.

\subsection{Ablation Studies}

\begin{table*}[!ht]
\centering
\caption{Ablation Studies on Velocity Prediction Model.}
\begin{tabular}{cccccccccccccc}
\hline
\rowcolor[HTML]{FFFFFF} 
\multicolumn{1}{c|}{\cellcolor[HTML]{FFFFFF}}                              & \multicolumn{3}{c|}{\cellcolor[HTML]{FFFFFF}}                                                                                                                                                                                                                  & \multicolumn{2}{c|}{\cellcolor[HTML]{FFFFFF}\textbf{NYUv2}}                                    & \multicolumn{2}{c|}{\cellcolor[HTML]{FFFFFF}\textbf{KITTI}}                                    & \multicolumn{2}{c|}{\cellcolor[HTML]{FFFFFF}\textbf{ETH3D}}                                    & \multicolumn{2}{c|}{\cellcolor[HTML]{FFFFFF}\textbf{ScanNet}}                                  & \multicolumn{2}{c}{\cellcolor[HTML]{FFFFFF}\textbf{DIODE}}  \\ \cline{5-14} 
\rowcolor[HTML]{FFFFFF} 
\multicolumn{1}{c|}{\cellcolor[HTML]{FFFFFF}}                              & \multicolumn{3}{c|}{\multirow{-2}{*}{\cellcolor[HTML]{FFFFFF}\textbf{Ablation Settings}}}                                                                                                                                                                      & \textbf{AR$\downarrow$}                 & \multicolumn{1}{c|}{\cellcolor[HTML]{FFFFFF}\textbf{$\delta_1\uparrow$}} & \textbf{AR$\downarrow$}                 & \multicolumn{1}{c|}{\cellcolor[HTML]{FFFFFF}\textbf{$\delta_1\uparrow$}} & \textbf{AR$\downarrow$}                 & \multicolumn{1}{c|}{\cellcolor[HTML]{FFFFFF}\textbf{$\delta_1\uparrow$}} & \textbf{AR$\downarrow$}                 & \multicolumn{1}{c|}{\cellcolor[HTML]{FFFFFF}\textbf{$\delta_1\uparrow$}} & \textbf{AR$\downarrow$}                  & \textbf{$\delta_1\uparrow$}          \\ \cline{2-14} 
\rowcolor[HTML]{D3FFDA} 
\multicolumn{1}{c|}{\cellcolor[HTML]{FFFFFF}}                              & \multicolumn{3}{c|}{\cellcolor[HTML]{FFFFFF}\textbf{Full Model}}                                                                                                                                                                                               & \textbf{4.0}                & \textbf{98.1}                                                    & \textbf{6.2}                & \textbf{95.9}                                                    & \textbf{3.9}                & \textbf{98.8}                                                    & \textbf{4.3}                & \textbf{97.6}                                                    & \textbf{18.8}                & \textbf{80.4}                \\ \cline{2-14} 
\rowcolor[HTML]{F2F2F2} 
\multicolumn{1}{c|}{\multirow{-4}{*}{\cellcolor[HTML]{FFFFFF}\textbf{\#}}} & \textbf{\begin{tabular}[c]{@{}c@{}}Semantic\\ Decoder\end{tabular}} & \textbf{\begin{tabular}[c]{@{}c@{}}Coarse\\ Decoder\end{tabular}} & \multicolumn{1}{c|}{\cellcolor[HTML]{F2F2F2}\textbf{\begin{tabular}[c]{@{}c@{}}Fine-grained\\ Decoder\end{tabular}}} & \multicolumn{10}{c}{\cellcolor[HTML]{F2F2F2}\textit{\textbf{Configurations of Decoders}}}                                                                                                                                                                                                                                                                                                                                                                       \\ \hline
\rowcolor[HTML]{FFFFFF} 
\multicolumn{1}{c|}{\cellcolor[HTML]{FFFFFF}\textbf{I}}                    & \textbf{4-Layer}                                                    & \textbf{8-Layer}                                                  & \multicolumn{1}{c|}{\cellcolor[HTML]{FFFFFF}\textbf{×}}                                                              & 5.8                         & \multicolumn{1}{c|}{\cellcolor[HTML]{FFFFFF}97.2}                & \cellcolor[HTML]{FFE8E8}8.3 & \multicolumn{1}{c|}{\cellcolor[HTML]{FFE8E8}94.0}                & 6.1                         & \multicolumn{1}{c|}{\cellcolor[HTML]{FFFFFF}97.4}                & 5.6                         & \multicolumn{1}{c|}{\cellcolor[HTML]{FFFFFF}96.8}                & 21.7                         & 78.7                         \\
\rowcolor[HTML]{FFFFFF} 
\multicolumn{1}{c|}{\cellcolor[HTML]{FFFFFF}\textbf{II}}                   & \textbf{4-Layer}                                                    & \textbf{×}                                                        & \multicolumn{1}{c|}{\cellcolor[HTML]{FFFFFF}\textbf{8-Layer}}                                                        & \cellcolor[HTML]{FFE8E8}5.5 & \multicolumn{1}{c|}{\cellcolor[HTML]{FEFFBB}97.5}                & 8.9                         & \multicolumn{1}{c|}{\cellcolor[HTML]{FFFFFF}93.8}                & \cellcolor[HTML]{FFE8E8}5.9 & \multicolumn{1}{c|}{\cellcolor[HTML]{FFFFFF}97.5}                & \cellcolor[HTML]{FEFFBB}5.3 & \multicolumn{1}{c|}{\cellcolor[HTML]{FEFFBB}97.2}                & \cellcolor[HTML]{FFE8E8}20.7 & 79.6                         \\
\rowcolor[HTML]{FFFFFF} 
\multicolumn{1}{c|}{\cellcolor[HTML]{FFFFFF}\textbf{III}}                  & \textbf{×}                                                          & \textbf{12-Layer}                                                 & \multicolumn{1}{c|}{\cellcolor[HTML]{FFFFFF}\textbf{×}}                                                              & 6.2                         & \multicolumn{1}{c|}{\cellcolor[HTML]{FFFFFF}96.8}                & 10.5                        & \multicolumn{1}{c|}{\cellcolor[HTML]{FFFFFF}89.6}                & 8.2                         & \multicolumn{1}{c|}{\cellcolor[HTML]{FFFFFF}95.7}                & 6.1                         & \multicolumn{1}{c|}{\cellcolor[HTML]{FFFFFF}96.3}                & 22.1                         & 77.9                         \\
\rowcolor[HTML]{FFFFFF} 
\multicolumn{1}{c|}{\cellcolor[HTML]{FFFFFF}\textbf{IV}}                   & \textbf{×}                                                          & \textbf{×}                                                        & \multicolumn{1}{c|}{\cellcolor[HTML]{FFFFFF}\textbf{12-Layer}}                                                       & \cellcolor[HTML]{FFE8E8}5.5 & \multicolumn{1}{c|}{\cellcolor[HTML]{FFE8E8}97.4}                & 9.8                         & \multicolumn{1}{c|}{\cellcolor[HTML]{FFFFFF}92.3}                & 6.9                         & \multicolumn{1}{c|}{\cellcolor[HTML]{FFFFFF}97.0}                & \cellcolor[HTML]{FFE8E8}5.4 & \multicolumn{1}{c|}{\cellcolor[HTML]{FFFFFF}96.9}                & 21.4                         & 79.0                         \\
\rowcolor[HTML]{FFFFFF} 
\multicolumn{1}{c|}{\cellcolor[HTML]{FFFFFF}\textbf{V}}                    & \textbf{×}                                                          & \textbf{6-Layer}                                                  & \multicolumn{1}{c|}{\cellcolor[HTML]{FFFFFF}\textbf{6-Layer}}                                                        & \cellcolor[HTML]{FFE8E8}5.5 & \multicolumn{1}{c|}{\cellcolor[HTML]{FFFFFF}97.3}                & 9.8                         & \multicolumn{1}{c|}{\cellcolor[HTML]{FFFFFF}92.5}                & 6.0                         & \multicolumn{1}{c|}{\cellcolor[HTML]{FFE8E8}97.6}                & \cellcolor[HTML]{FEFFBB}5.3 & \multicolumn{1}{c|}{\cellcolor[HTML]{FFFFFF}96.8}                & \cellcolor[HTML]{FEFFBB}20.5 & \cellcolor[HTML]{FEFFBB}79.8 \\
\rowcolor[HTML]{FFFFFF} 
\multicolumn{1}{c|}{\cellcolor[HTML]{FFFFFF}\textbf{VI}}                   & \textbf{4-Layer}                                                    & \textbf{6-Layer}                                                  & \multicolumn{1}{c|}{\cellcolor[HTML]{FFFFFF}\textbf{2-Layer}}                                                        & 6.0                         & \multicolumn{1}{c|}{\cellcolor[HTML]{FFFFFF}96.9}                & 8.8                         & \multicolumn{1}{c|}{\cellcolor[HTML]{FFFFFF}93.7}                & 6.3                         & \multicolumn{1}{c|}{\cellcolor[HTML]{FFFFFF}97.3}                & 6.0                         & \multicolumn{1}{c|}{\cellcolor[HTML]{FFFFFF}96.2}                & 21.6                         & 78.6                         \\
\rowcolor[HTML]{FFFFFF} 
\multicolumn{1}{c|}{\cellcolor[HTML]{FFFFFF}\textbf{VII}}                  & \textbf{2-Layer}                                                    & \textbf{6-Layer}                                                  & \multicolumn{1}{c|}{\cellcolor[HTML]{FFFFFF}\textbf{4-Layer}}                                                        & 6.0                         & \multicolumn{1}{c|}{\cellcolor[HTML]{FFFFFF}96.8}                & 9.3                         & \multicolumn{1}{c|}{\cellcolor[HTML]{FFFFFF}92.8}                & 6.6                         & \multicolumn{1}{c|}{\cellcolor[HTML]{FFFFFF}97.1}                & 5.8                         & \multicolumn{1}{c|}{\cellcolor[HTML]{FFFFFF}96.6}                & 22.5                         & 77.9                         \\ \hline
\rowcolor[HTML]{F2F2F2} 
\textbf{}                                                                  & \multicolumn{13}{c}{\cellcolor[HTML]{F2F2F2}\textit{\textbf{Injection of Semantic Feature}}}                                                                                                                                                                                                                                                                                                                                                                                                                                                                                                                                                                                                                                     \\
\rowcolor[HTML]{FFFFFF} 
\multicolumn{1}{c|}{\cellcolor[HTML]{FFFFFF}\textbf{VIII}}                 & \multicolumn{3}{c|}{\cellcolor[HTML]{FFFFFF}\textbf{Cross Attention}}                                                                                                                                                                                          & 6.3                         & \multicolumn{1}{c|}{\cellcolor[HTML]{FFFFFF}96.6}                & 34.2                        & \multicolumn{1}{c|}{\cellcolor[HTML]{FFFFFF}42.5}                & 8.2                         & \multicolumn{1}{c|}{\cellcolor[HTML]{FFFFFF}94.5}                & 6.0                         & \multicolumn{1}{c|}{\cellcolor[HTML]{FFFFFF}96.5}                & 23.5                         & 75.6                         \\
\rowcolor[HTML]{FEFFBB} 
\multicolumn{1}{c|}{\cellcolor[HTML]{FFFFFF}\textbf{IX}}                   & \multicolumn{3}{c|}{\cellcolor[HTML]{FFFFFF}\textbf{Concatenation}}                                                                                                                                                                                            & 5.2                         & \multicolumn{1}{c|}{\cellcolor[HTML]{FEFFBB}97.5}                & 8.2                         & \multicolumn{1}{c|}{\cellcolor[HTML]{FEFFBB}94.1}                & 5.5                         & \multicolumn{1}{c|}{\cellcolor[HTML]{FEFFBB}97.9}                & 5.3                         & \multicolumn{1}{c|}{\cellcolor[HTML]{FFE8E8}97.1}                & \cellcolor[HTML]{FFE8E8}20.7 & \cellcolor[HTML]{FFFFFF}79.6 \\
\rowcolor[HTML]{FFFFFF} 
\multicolumn{1}{c|}{\cellcolor[HTML]{FFFFFF}\textbf{X}}                    & \multicolumn{3}{c|}{\cellcolor[HTML]{FFFFFF}\textbf{AdaNorm}}                                                                                                                                                                                                  & \cellcolor[HTML]{FFE8E8}5.5 & \multicolumn{1}{c|}{\cellcolor[HTML]{FFFFFF}97.3}                & 9.2                         & \multicolumn{1}{c|}{\cellcolor[HTML]{FFFFFF}93.2}                & \cellcolor[HTML]{FFE8E8}5.9 & \multicolumn{1}{c|}{\cellcolor[HTML]{FFFFFF}97.3}                & \cellcolor[HTML]{FFE8E8}5.4 & \multicolumn{1}{c|}{\cellcolor[HTML]{FFFFFF}97.0}                & 20.8                         & \cellcolor[HTML]{FFE8E8}79.7 \\
\rowcolor[HTML]{FFFFFF} 
\multicolumn{1}{c|}{\cellcolor[HTML]{FFFFFF}\textbf{XI}}                   & \multicolumn{3}{c|}{\cellcolor[HTML]{FFFFFF}\textbf{No Semantic}}                                                                                                                                                                                              & 20.6                        & \multicolumn{1}{c|}{\cellcolor[HTML]{FFFFFF}65.5}                & 26.0                        & \multicolumn{1}{c|}{\cellcolor[HTML]{FFFFFF}56.9}                & 20.1                        & \multicolumn{1}{c|}{\cellcolor[HTML]{FFFFFF}69.2}                & 18.7                        & \multicolumn{1}{c|}{\cellcolor[HTML]{FFFFFF}69.9}                & 32.6                         & 59.1                         \\ \hline
\end{tabular}
\label{ab1}
\end{table*}

\begin{figure}
    \centering
    \includegraphics[width=\linewidth]{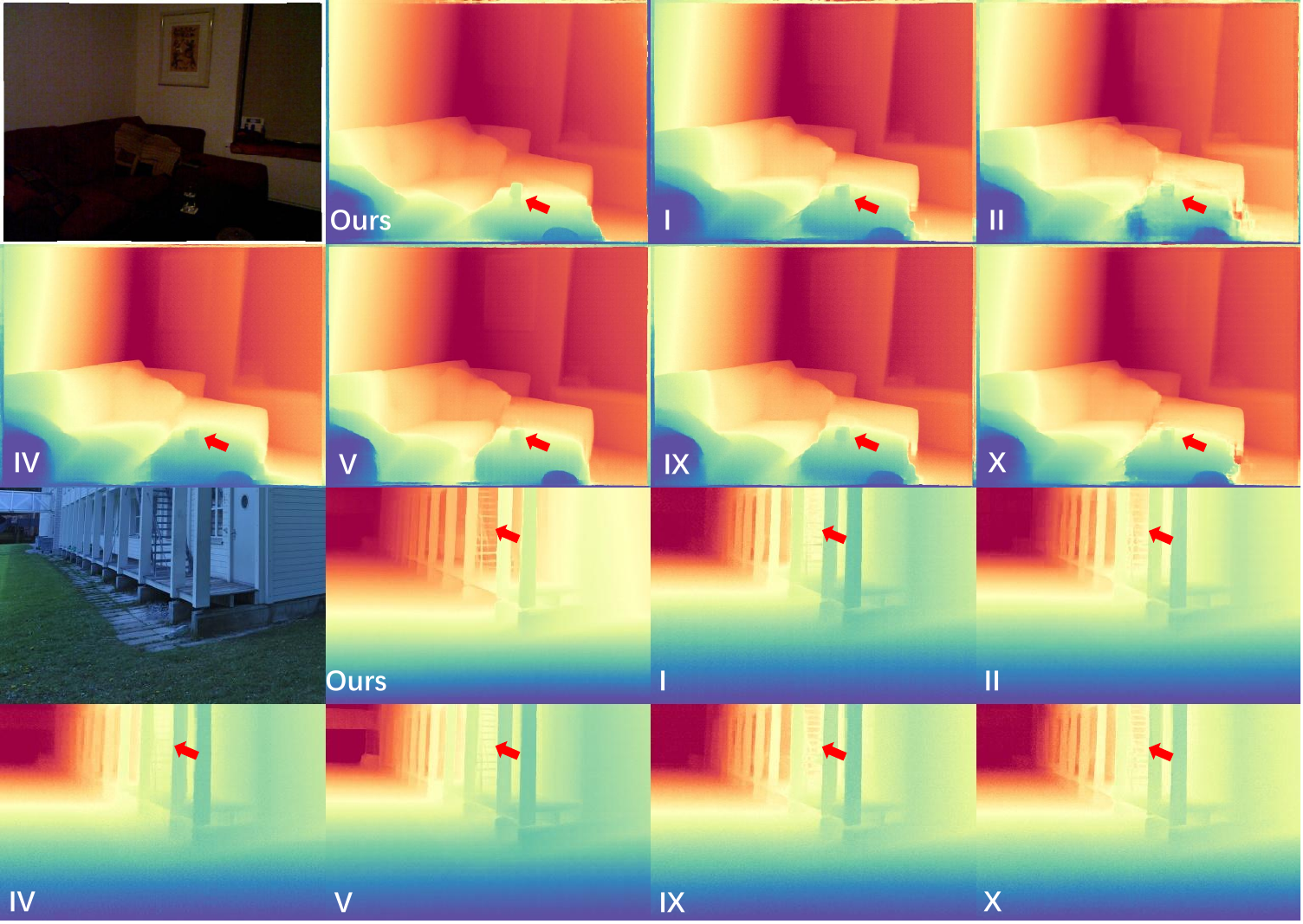}
    \caption{Qualitative comparison of ablation settings.}
    \label{ab2}
\end{figure}

Firstly, we perform ablations on velocity prediction model by re-organizing semantic decoder, coarse decoder and fine-grained decoder with different numbers of layers. Meanwhile, we also use different injection methods including cross-attention, concatenation and adaptive normalization instead of our modulation to perform ablation studies. To control the computational cost of models, we keep the overall number of decoder layers as 12. According to results shown in Table \ref{ab1}. Besides, we select several ablation settings with best performance to perform qualitative comparison in Fig. \ref{ab2} and compare point cloud reconstruction with our full model in Table. \ref{ab3}. Furthermore, we respectively conduct 1-step, 2-step, 4-step, 6-step and 8-step inference to demonstrate the influence of flow matching steps, results of which are displayed in Table \ref{ab4}. We can find that 2-step inference via 4-layer decoders with the propsoed semantic injection is the best choice for monocular depth estimation.

\begin{table}[]
\centering
\caption{Ablation results on point cloud reconstruction. \textbf{Best models} are in bold.}
\begin{tabular}{c|cc|cc|cc}
\hline
\rowcolor[HTML]{FFFFFF} 
\cellcolor[HTML]{FFFFFF}                                    & \multicolumn{2}{c|}{\cellcolor[HTML]{FFFFFF}\textbf{NYUv2}} & \multicolumn{2}{c}{\cellcolor[HTML]{FFFFFF}\textbf{ETH3D}}                   & \multicolumn{2}{c|}{\cellcolor[HTML]{FFFFFF}\textbf{Mean}} \\ \cline{2-7}
\rowcolor[HTML]{FFFFFF} 
\multirow{-2}{*}{\cellcolor[HTML]{FFFFFF}\textbf{Ablation}} & \textbf{Acc↓}                & \textbf{CD↓}                 & \textbf{Acc↓}  & \multicolumn{1}{c|}{\cellcolor[HTML]{FFFFFF}\textbf{CD↓}}   & \textbf{Acc↓}               & \textbf{CD↓}                \\ \hline
\rowcolor[HTML]{FFFFFF} 
\multicolumn{1}{c|}{\cellcolor[HTML]{FFFFFF}\textbf{II}}    & 118.0                        & 119.9                        & 455.4          & \multicolumn{1}{c|}{\cellcolor[HTML]{FFFFFF}383.1}          & 286.7                       & 251.5                       \\
\rowcolor[HTML]{FFFFFF} 
\multicolumn{1}{c|}{\cellcolor[HTML]{FFFFFF}\textbf{IV}}    & 132.2                        & 123.1                        & 460.2          & \multicolumn{1}{c|}{\cellcolor[HTML]{FFFFFF}368.1}          & 296.2                       & 245.6                       \\
\rowcolor[HTML]{FFFFFF} 
\multicolumn{1}{c|}{\cellcolor[HTML]{FFFFFF}\textbf{V}}     & 135.2                        & 128.5                        & 902.5          & \multicolumn{1}{c|}{\cellcolor[HTML]{FFFFFF}621.1}          & 518.8                       & 374.8                       \\
\rowcolor[HTML]{FFFFFF} 
\multicolumn{1}{c|}{\cellcolor[HTML]{FFFFFF}\textbf{Ours}}  & \textbf{89.3}                & \textbf{100.2}               & \textbf{269.6} & \multicolumn{1}{c|}{\cellcolor[HTML]{FFFFFF}\textbf{299.1}} & \textbf{179.5}              & \textbf{199.7}              \\ \hline
\end{tabular}
\label{ab3}
\end{table}

\begin{table}
\centering
\caption{Ablation studies on flow matching steps. \textbf{Best models} are in bold, while \underline{Second Best models} are underlined.}
\fontsize{6pt}{8pt}\selectfont
\begin{tabular}{c|cc|cc|cc|cc|cc}
\hline
\rowcolor[HTML]{FFFFFF} 
\cellcolor[HTML]{FFFFFF}                                 & \multicolumn{2}{c|}{\cellcolor[HTML]{FFFFFF}\textbf{NYUv2}} & \multicolumn{2}{c|}{\cellcolor[HTML]{FFFFFF}\textbf{KITTI}} & \multicolumn{2}{c|}{\cellcolor[HTML]{FFFFFF}\textbf{ETH3D}} & \multicolumn{2}{c|}{\cellcolor[HTML]{FFFFFF}\textbf{ScanNet}} & \multicolumn{2}{c}{\cellcolor[HTML]{FFFFFF}\textbf{DIODE}} \\ \cline{2-11} 
\rowcolor[HTML]{FFFFFF} 
\multirow{-2}{*}{\cellcolor[HTML]{FFFFFF}\textbf{Steps}} & \textbf{AR}               & \textbf{$\delta_1$}             & \textbf{AR}               & \textbf{$\delta_1$}             & \textbf{AR}               & \textbf{$\delta_1$}             & \textbf{AR}                & \textbf{$\delta_1$}              & \textbf{AR}               & \textbf{$\delta_1$}            \\ \hline
\rowcolor[HTML]{FFFFFF} 
\textbf{1}                                               & 4.4                       & \textbf{98.1}                   & 7.5                       & 94.7                            & 4.6                       & 98.5                            & 4.5                        & \textbf{97.6}                    & \textbf{18.8}             & {\ul
 80.3}                     \\
\rowcolor[HTML]{EFEFEF} 
\textbf{2}                                               & \textbf{4.0}              & \textbf{98.1}                   & \textbf{6.2}              & \textbf{95.9}                   & \textbf{3.9}              & \textbf{98.8}                   & \textbf{4.3}               & \textbf{97.6}                    & \textbf{18.8}             & \textbf{80.4}
                  \\
\rowcolor[HTML]{FFFFFF} 
\textbf{4}                                               & {\ul 4.1}                 & {\ul 98.0}                      & {\ul 6.5}                 & {\ul 95.6}                      & {\ul 4.0}                 & {\ul 98.7}                      & \textbf{4.3}               & {\ul 97.5}                       & {\ul 18.9}                & {\ul
 80.3}                     \\
\rowcolor[HTML]{FFFFFF} 
\textbf{6}                                               & 4.2                       & {\ul 98.0}                      & 6.7                       & 95.4                            & 4.1                       & {\ul 98.7}                      & {\ul 4.4}                  & {\ul
 97.5}                       & 19.0                      & 80.2                           \\
\rowcolor[HTML]{FFFFFF} 
\textbf{8}                                               & 4.2                       & {\ul 98.0}                      & 6.7                       & 95.4                            & 4.1                       & {\ul 98.7}                      & {\ul 4.4}                  & {\ul 97.5}                       & 19.0                      & 80.2                           \\ \hline
\end{tabular}
\label{ab4}
\end{table}

\begin{table}[!ht]
\centering
\caption{Comparison of Computational Cost and Efficiency}
\begin{tabular}{l|cccc}
\hline
\rowcolor[HTML]{FFFFFF} 
\multicolumn{1}{c|}{\cellcolor[HTML]{FFFFFF}\textbf{Methods}} & \textbf{GFLOPs} & \textbf{ms/image} & \textbf{\begin{tabular}[c]{@{}c@{}}Mem\\ (GB)\end{tabular}} & \textbf{\begin{tabular}[c]{@{}c@{}}Params\\ (M)\end{tabular}} \\ \hline
\rowcolor[HTML]{FFFFFF} 
\textbf{UniDepth   v2}                                        & 1925.0          & 23.3              & 3.98                                                        & 353.8                                                         \\
\rowcolor[HTML]{FFFFFF} 
\textbf{MoGe}                                                 & 3108.0          & 56.8              & 3.65                                                        & 314.2                                                         \\
\rowcolor[HTML]{FFFFFF} 
\textbf{DepthMaster}                                          & 5241.5          & 94.7              & 6.71                                                        & 1028.2                                                        \\
\rowcolor[HTML]{FFFFFF} 
\textbf{PPD}                                                  & 18158.8         & 193.7             & 4.22                                                        & 804.2                                                         \\
\rowcolor[HTML]{EFEFEF} 
\textbf{FlowMoDE}                                             & \textbf{3140.1} & \textbf{53.4}     & \textbf{1.87}                                               & \textbf{398.7}                                                \\ \hline
\end{tabular}
\label{cost}
\end{table}

\subsection{Computational Cost}
We compare the computational cost of our model with several existing models with relatively best performance. We use GPU memory requirement (Mem) and number of parameters (Params) to evaluate the model size. Meanwhile, GFLOPs and inference speed (ms per image) are utilized to evaluate the efficiency of models. As shown in Table \ref{cost}, the proposed FlowMoDE incurs higher latency and model weight overhead compared with UniDepth v2, but achieves significantly lower computational cost than prior generative depth models.

\section{CONCLUSIONS}
In this paper, we present FlowMoDE, a flow‑matching generative model for sim‑to‑real monocular depth estimation to recover fine structural details. Trained solely on synthetic data, our model outperforms prior methods on five diverse benchmarks. We further validate its strong ability to reconstruct fine‑grained geometry in various robotic scenarios against state‑of‑the‑art approaches. With fewer Transformer blocks and minimal inference steps, FlowMoDE is lighter and more efficient than existing generative depth models. Still, patch embedding in our flow‑matching pipeline may produce occasional flying pixels. Moreover, training only on near‑field indoor synthetic data combined with patch‑wise denoising can cause blurred edges for distant regions.

\bibliographystyle{IEEEtran}
\bibliography{refs}

\end{document}